\documentclass[letterpaper,twocolumn,10pt]{article}
\usepackage{nopageno}
\usepackage{usenix}

\usepackage{tikz}
\usepackage{float}
\usepackage{amsmath}
\usepackage{balance}
\usepackage{booktabs}
\usepackage[capitalise]{cleveref}
\usepackage{xcolor}
\usepackage{xspace}
\usepackage{multirow}
\usepackage{tabularx}
\usepackage{algorithm}
\usepackage{multicol}
\usepackage{algpseudocode}
\usepackage{enumitem}
\usepackage{makecell}
\algtext*{EndWhile}
\algtext*{EndIf}
\algtext*{EndFor}
\usepackage{caption}
\usepackage{subcaption}
\usepackage[compact]{titlesec}
\usepackage{fancyhdr}
\usepackage{enumitem}
\setitemize{noitemsep,topsep=0pt,parsep=0pt,partopsep=0pt}

\newif\ifcomments
\commentstrue 
\ifcomments
    
\else
    
\fi

\newcommand{\sys}[0]{Step-3\xspace}
\def\Snospace~{\S{}}

\newcommand{\parabf}[1]{\medskip\noindent\textbf{#1}}

\newcommand{\cut}[1]{}

\begin{document}

\date{}

\title{\sys is Large yet Affordable: \\ Model-system Co-design for Cost-effective Decoding}

\author{
\rm{StepFun Inc.}
}

\maketitle

\begin{abstract}
Large language models (LLMs) face low hardware efficiency during decoding, especially for long-context reasoning tasks. This paper introduces Step-3, a 321B-parameter VLM with hardware-aware model-system co-design optimized for minimizing decoding costs. Step-3 innovates in two key dimensions: (1) A novel Multi-Matrix Factorization Attention (MFA) mechanism that significantly reduces \emph{both} KV cache size and computation while maintaining high attention expressiveness, and (2) Attention-FFN Disaggregation (AFD), a distributed inference system that decouples attention and Feed-Forward Network (FFN) layers into specialized subsystems. This co-design achieves unprecedented cost efficiency: Step-3 significantly reduces theoretical decoding costs compared with models like DeepSeek-V3 and Qwen3 MoE 235B, with the gains widening at longer context. 
Step-3 achieves low cost while activating 38B parameters per token (more than DeepSeek-V3 and Qwen3 MoE 235B), demonstrating that hardware-aligned attention arithmetic intensity, MoE sparsity, and AFD are critical to cost-effectiveness. We perform a head-to-head comparison with DeepSeek-V3 in its favorable scenarios. Our implementation on Hopper GPUs achieves a decoding throughput of up to 4,039 tokens per second per GPU under 50ms TPOT SLA (4K context, FP8, no MTP). It is higher than DeepSeek-V3's 2,324 in the same setup and sets a new Pareto frontier for LLM decoding.

\end{abstract}

\section{Introduction}
\label{sec:introduction}

This paper presents the model-system co-design of \sys, specifically engineered for the test-time scaling paradigm with the primary optimization objective of minimizing decoding costs. \sys has 321 billion total parameters, while for each text token, 38B parameters are activated. We will demonstrate that, although \sys is in the multi-hundred billion parameter range and the activated parameters are slightly larger than representative open-weight models like DeepSeek V3 (DSv3)~\cite{deepseekv3}, we achieve significantly lower decoding costs with model-system co-design.

We focus on optimizing decoding because 1) it is the most expensive per token (because of low MFU) compared with training and prefill. 2) For reasoning models, longer thinking leads to higher intelligence, so lowering decoding costs can translate to higher intelligence for fixed-budget scenarios. 3) Faster and cheaper decoding also speeds up the RL training. 4) There is a large room for optimization and is therefore more technically interesting.


Recently, there emerged several large open-weight models.
Some of them explored novel architecture changes on top of traditional Transformers. The innovations focus on the two main Transformer components – there are new attention designs to reduce KV cache overhead during inference, and there are Mixture-of-Experts (MoE) structures to enhance FFN while limiting the growth of computation requirements.

\begin{figure}[!t]
    \centering
    \includegraphics[width=0.95\linewidth]{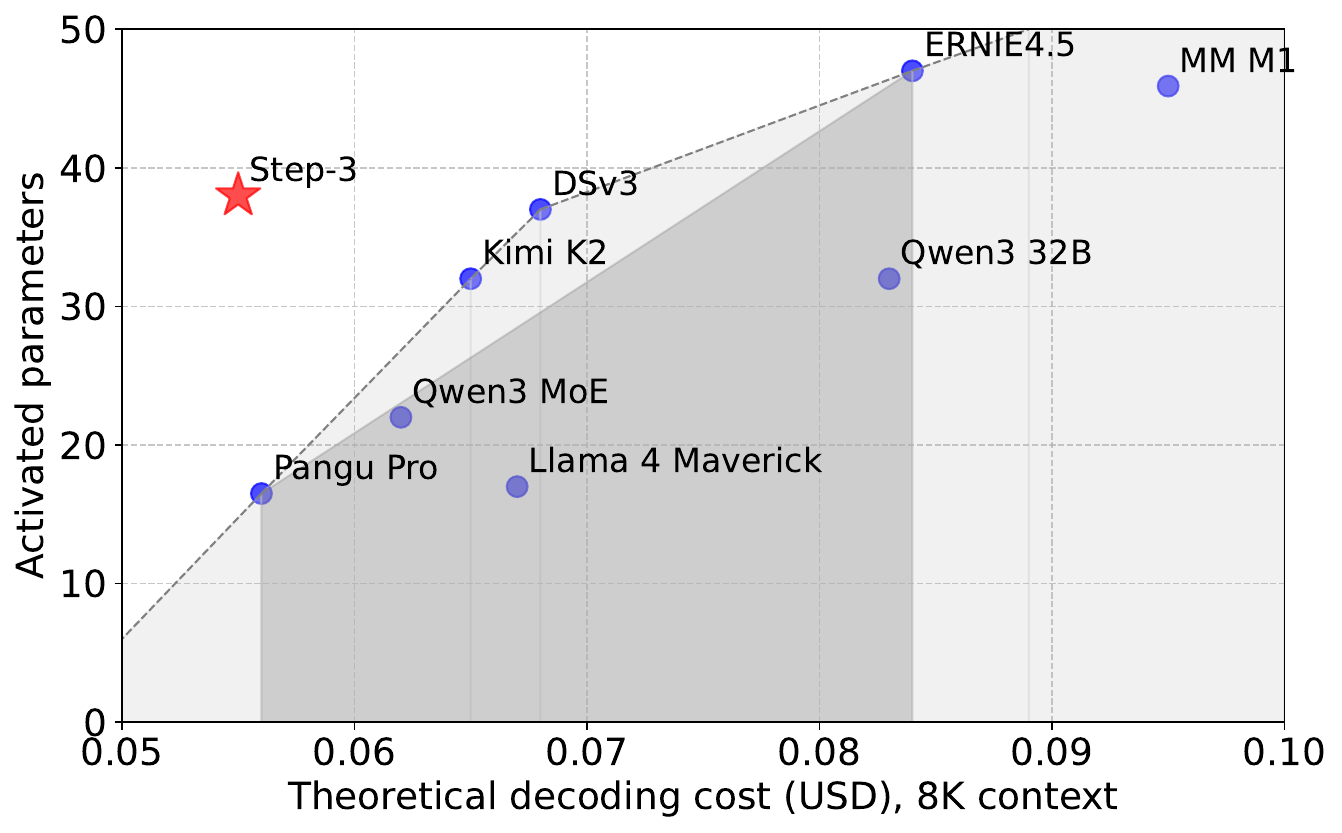}
    \caption{The Pareto frontier of recent models regarding activated parameters and decoding costs. The darker area is GQA models' Pareto frontier. Note: \sys also has the highest attention effective rank~\cite{hu2025multimatrixfactorizationattention}, the same as DSv3 and doubling some other models like Qwen3 MoE 235B and Kimi K2.}
    \label{fig:cost_pareto}
\end{figure}


We also started to work on model architecture exploration, \emph{e.g.,} through MoE model development (Step-2~\cite{step2}) since late 2023 and MFA~\cite{hu2025multimatrixfactorizationattention}, a new attention architecture released in late 2024. In the process, observing the recent open-weight models, we identify two common suboptimal practices:

\begin{itemize}[leftmargin=*]
  \item For attention, some are overly emphasizing on reducing KV cache sizes, at excessive cost of computation load. It makes the model less cost-effective to run on more affordable but weaker hardware. Meanwhile, it limits the room for other acceleration techniques like quantization and speculative decoding.
  \item For FFN, some are overly emphasizing on pursuing sparser architectures without considering whether they fit today's hardware. It either harms the hardware efficiency, or lowering model performance without gaining cost advantages.
\end{itemize}

Hoping to inspire more discussion and rethinking about the above trends, we report our recent progress, \sys, and the analysis and rationale behind its design. The outcome is promising -- in Figure~\ref{fig:cost_pareto}, we show the best theoretical decoding costs of \sys and recent models. For each model, we searched for the best deployment strategy based on Attention-FFN Disaggregation (AFD, \S\ref{sec:af}), which we advocate, and any combination of H800, H20, A800 or Ascend 910B.\footnote{In practice, the FFN part of DSv3, Kimi K2 and Llama 4 Maverick may suffer from MoE over-sparsity (\S\ref{subsec:moe_sparsity}) and be farther from theoretical costs. In Figure~\ref{fig:cost_pareto}, we give them a favor by ignoring the issue.} \sys largely improves the Pareto frontier of activated parameters and decoding costs. Though not shown in the figure, its advantage continues to widen with longer context.\footnote{Some may wonder about hybrid linear attention models like MiniMax M1. We will discuss more in \S\ref{subsec:demystify}.}


Our work is based on the assumption of deploying prior work of Prefill-Decoding (PD) disaggregation~\cite{distserve,patel2023splitwise}. With it, we can focus only on optimizing decoding, without worrying about the impact on prefill. Readers will see similar benefits of deploying AFD, \emph{i.e.,} how it allows us to divide-and-conquer attention and FFN designs. It leads to a model architecture whose both parts are more cost-effective. We implement the inference system and show that \sys indeed achieves much lower decoding costs compared with other multi-billion parameter models.


Below is a summary of our findings.
\begin{itemize}[leftmargin=*]
  \item 
  \textbf{Decoding costs go beyond parameter count:} Neither the total parameter count or activated parameter count is a good indicator for decoding costs.
  \begin{itemize}[leftmargin=*]
    \item For example, Qwen-3 MoE 235B exhibits only $10\%$ lower theoretical decoding cost (on H20, the best hardware for it) than DSv3 (on H800, the best hardware for DSv3) despite having $65\%$ fewer total parameters and $40\%$ fewer activated parameters.
    
    \item \sys achieves $\sim40\%$ decoding cost reduction versus both models despite its total parameter count being between the two models and having the highest activation parameters.
  \end{itemize}

  \item \textbf{The attention design dominates decoding costs:} With AFD, we decouple the cost analysis of attention and FFN because we can run them in the most cost-effective way, respectively. Then it becomes apparent that the attention design has a larger impact on decoding costs than (total or activated) parameter count.

  \item \textbf{KV cache size is not the single factor impacting attention costs:} We find that some attention designs requires too much computation (too high arithmetic intensity) for lower-cost hardware platforms. More importantly, we are the first to show this problem indeed affects the final decoding costs and thus leaves large room for \sys to achieve significant cost savings.
  

  \item \textbf{MoE needs hardware-aware design:} The degree of MoE sparsity must joinly consider hardware's computation power, memory bandwidth and network bandwidth. Overly sparse models may have small activated parameters on paper, but run inefficiently on today's hardware.

  \item \textbf{For decoding acceleration, the devil is in the details:} Linear attention, quantization, and MTP are all promising directions to accelerate decoding. However, some design points that may seem nuance can remove most of the benefits in decoding. 

  \item \textbf{AFD deployment:} We believe it is the superior decoding system design compared with existing solutions, because of the following unique advantages:
  \begin{itemize}[leftmargin=*]
    \item Facilitating divide-and-conquer model design.
    \item Easy scaling of attention instances to handle dynamic context length.
    \item Always keeping an ideal batch size for FFN to achieve high MFU, independent from  attention.
    \item Overlapping communication overhead with a perfectly balanced pipeline.
    \item Reducing the scale requirement compared with DeepEP~\cite{deepep2025}, and getting better reliability and less EP imbalance.
    \item Allowing the use of heterogeneous hardware to further reduce decoding costs.
  \end{itemize}
\end{itemize}

\section{\sys Model Card}
\label{sec:model_card}

Before diving into the model-system co-design details, we briefly describe \sys.

\sys is built upon the Transformer architecture \cite{vaswani2017attention}, with each Transformer block comprising an attention module and a Feed-Forward Network (FFN). For the attention mechanism, we introduce Multi-Matrix Factorization Attention (MFA) \cite{hu2025multimatrixfactorizationattention}, which leverages low-rank matrix factorization in the Query-Key (QK) circuit \cite{elhage2021mathematical}. This design enables parameter-efficient scaling of both the number and dimensionality of attention heads while minimizing KV cache overhead. For FFNs, we adopt a shared expert design inspired by DeepSeekMoE, incorporating Mixture-of-Experts (MoE) layers. Our configuration includes 61 Transformer layers with a hidden dimension of 7168. 
For MFA, we configure 64 query heads and they share a Key and a Value head, all with a dimension of 256. The query dimension is down-projected from 7168 to a lower-rank of 2048, followed by a normalization, and then up-projected to 64*256. 
MoE layers are applied to all FFNs except the first four and the last layer. Under this setup, \sys comprises 316 billion parameters, with 38 billion activated per token. There is an additional vision encoder of 5 billion parameters, which we do not discuss in this paper because it is irrelevant to decoding.

In the future, we will release more details on the model side for \sys.

\begin{table}[htbp]
\centering

\renewcommand{\arraystretch}{1.2}
\begin{tabular}{|l|l|}
\hline \textbf{} & \textbf{\sys}      \\ \hline
{\# Layers} & 61 \\ \hline
{Hidden Dimension} & 7168 \\ \hline
{Attention Mechanism} & MFA \\ \hline
{Low-rank Query Dimension} & 2048 \\ \hline
{\# Query Heads} & 64 \\ \hline
{Head Dimension} & 256 \\ \hline
{\# Shared Experts} & 1 \\ \hline
{MoE Layer Configuration} & All layers except the \\ 
& first four and last layer \\ \hline
{Total Parameters (LLM)} & 316 Billion \\ \hline
{Activated Params per Token} & 38 Billion \\ \hline
{Total Parameters (VLM)} & 321 Billion \\ \hline

\end{tabular}

\caption{Model card for \sys.}
\label{tab:model-card-step3}
\end{table}

\section{Attention-FFN Disaggregation}
\label{sec:af}

We start by describing \sys inference system, which may be one of the first production quality serving systems that leverages the Attention-FFN Disaggregation (AFD) idea and achieves high-throughput decoding under strict SLO constraints. First, we elaborate on the rationale behind the AFD design.

\parabf{Rationale.}
LLMs are typically composed of interleaved attention and Feed-Forward Network (FFN) layers, each exhibiting distinct computational and memory access patterns. 
For example, attention layers typically have a smaller number of parameters, but require storing the key-value cache (KV-cache) for each token, which is memory-intensive during inference. In contrast, FFN layers generally take up a much larger parameter count, especially for MoE models, yet do not require storing intermediate computation results. We will dive into the operational characteristics and inference costs of attention and FFN layers in \S\ref{sec:cost}.

Existing serving systems often treat these layers as monolithic blocks and overlook their intrinsic differences, leading to suboptimal GPU utilization. Hence, by disaggregating the attention and FFN components, we can better exploit their respective hardware affinities and optimize throughput. In addition, the disaggregation provides us with an opportunity to make an assumption: Both the attention and FFN parts can operate under ideal hardware conditions and can achieve high MFU, respectively. 

This idea is based on the Prefill-Decoding (PD) disaggregation approach~\cite{distserve}, which advocates separating the prefill and decoding stages to optimize resource utilization. 
Hence, we focus on the decoding stage with AFD without worrying about the impact on prefill. The analysis will become much simpler with this divide-and-conquer approach.


\subsection{Design Goals}

AFD deploys the attention and FFN layers onto separate sets of GPUs. 
This architectural separation allows each subsystem to adopt different parallelism strategies that best suit their computational characteristics.
During layer-wise decoding, hidden states are transmitted between the attention and FFN subsystems through high-speed network communication. 
This interleaved communication pattern forms a tightly coupled pipeline, where attention and FFN act as upstream and downstream stages for each other. 

Furthermore, network transmission latency must also be taken into account. In such fine-grained scenarios, its magnitude is comparable to the computation time of both attention and FFN stages. This means that the communication stage should also be considered when orchestrating the pipeline.

To achieve optimal overall performance, the processing latency of both sides must be precisely matched; any imbalance leads to pipeline stalls or under-utilized resources. 
Therefore, it is essential to jointly orchestrate the performance of A/F and communication stages.



We summarize the design goals of AFD as follows, which will be discussed in detail later:

\begin{itemize}[leftmargin=*]
    \item Performance target: 50ms time per output token (TPOT, $\ge 20$ tokens/sec) via a 3-stage pipeline, with 16.6ms per stage for A/F/communication, respectively. Here the time is accumulated across all model layers.\footnote{Alternatively, we can also use a 4-stage pipeline: A -> communication -> F -> communication, with a 12.5ms budget for each stage.}
    
    \item Pipeline optimization: Resource allocation and performance tuning that enable perfect A/F/communication multi-stages pipelining, hiding communication latency. 

    \item Independent design of A/F: With AFD, we can independently analyze the operational characteristics of attention and FFN. This separation not only enables optimal optimization for each subsystem, but also allows for flexible architectural modifications to the model itself. 
    
    \item Hardware selection: Independent hardware selection for attention and FFN subsystems based on their operational characteristics.

\end{itemize}


\subsection{Comparisons with Related Work}

\parabf{DeepSeek EP.} Large Expert Parallelism (EP) architecture is introduced in DeepSeek-V3~\cite{deepseekv3} to improve serving efficiency. 
Although EP also facilitates batch size amplification by distributing expert weights to multiple devices, we argue that this approach exhibits fundamental limitations compared to AFD.

\begin{itemize}[leftmargin=*]

\item {Deployment scale}:
A key advantage of AFD is its ability to operate efficiently at a smaller deployment scale. As mentioned before, DSv3 requires 320 GPUs for a decoding instance, while \sys only uses 32 GPUs (\S\ref{subsec:perf}). 
If the deployment scale expands significantly, network congestion becomes a critical issue~\cite{deepseek-infra}, resulting in increased and unpredictable latency. This heightened latency can severely impact the serving system's ability to meet inference SLA.

\item {Context-length efficiency}:
Long-context processing disproportionately burdens EP's attention layers, causing FFN under-utilized due to fixed expert-node allocation. AFD resolves this via decoupled scaling of attention and FFN. We will present quantitative results on different context lengths in \S\ref{sec:cost}.

\item {Load imbalance issue}: EP suffers from the well-known workload imbalanced issue~\cite{sigcomm23_janus, atc23_lina}. DeepSeek-V3 alleviates this issue using duplicated experts that can balance each GPU's workload in an ad-hoc manner. But this approach incurs additional memory overhead, and is inflexible to dynamic workload changes, especially when the data distribution shifts significantly. 
On the other hand, AFD can easily leverage hybrid TP-EP strategy to strike a balance between computation efficiency, communication traffic, and load balancing.

\item {Heterogeneous hardware constraints}:
AFD enables more flexible hardware deployment, in that attention and FFN instances can be mapped to heterogeneous hardware tailored to their respective compute and memory requirements, while EP forces homogeneous hardware deployment, limiting specialization benefits.

\item {Performance modeling}:
Our following analytical framework leverages the architectural disaggregation of attention and FFN. This separation provides methodological clarity due to their divergent computational profiles, which enables more accurate modeling of performance ceilings while substantially narrowing the gap between theoretical projections and empirical measurements. Contrarily, EP-only architecture lacks this divide-and-conquer clarity, suffering from inherent analytical ambiguity when modeling coupled subsystems.



\end{itemize}

In particular, we note that AFD is not a replacement for EP, but rather a complementary approach. In fact, \sys can be combined with the TP-EP strategy to achieve better performance and cost-effectiveness. \textit{The above analysis is against the EP-only architecture that does not employ AFD}, which is commonly used in existing serving systems~\cite{deepseekv3, ds_cloudmatrix384}.

\parabf{Megascale-Infer.}
To our knowledge, Megascale-Infer~\cite{megascale-infer} is the first to build a disaggregated serving system leveraging the AFD idea. However, it focuses on high throughput rather than providing a practical implementation to achieve the low latency target (\textit{i.e.,} 50ms TPOT) simultaneously. In fact, according to \cite{megascale-infer}, the reported latency per token of Megascale-Infer is 150ms, which is significantly higher than ours. Such high latency is not applicable for real-time applications like chatbots.
Moreover, the core of \sys is in model-system co-design, and we use the AFD idea to design \sys's model architecture for attention and FFN layers, while Megascale-Infer primarily only focuses on system-level optimizations.
We believe the co-design brings more opportunities to thoroughly exploit the hardware capabilities.
 
\section{Cost Analysis for LLM Decoding}
\label{sec:cost}

Given the important assumption that, with AFD, the attention part and the FFN part can operate near hardware limitations, we will now delve into the theoretical costs of each model. We compare \sys with several recently released models, namely DSv3~\cite{deepseekv3}, Kimi K2~\cite{kimik2}, Qwen3-235B-A22B~\cite{qwen3moe} (Qwen3-MoE for brevity), Qwen3-32B~\cite{qwen3}, Llama 4 Maverick~\cite{llama4}, MiniMax M1~\cite{minimaxm1} (MM M1), ERNIE 4.5~\cite{ernie4.5}, and Pangu Pro MoE~\cite{pangupromoe}.

\begin{table*}[h]
    \centering
    \begin{tabular}{l|p{2.7cm}|p{3.5cm}|p{3cm}|p{2.5cm}}
        \toprule
        Model & KV/State Memory Access (bytes) & Attention Computation w/o Linear (FLOPs) & Linear before and after Attention (FLOPs) & FFN Computation (FLOPs) \\
        \midrule
        DSv3 & $2.88 \times 10^8$ & $1.47 \times 10^{11}$ & $2.28 \times 10^{10}$ & $4.84 \times 10^{10}$ \\
        Kimi K2 & $2.88 \times 10^8$ & $7.37 \times 10^{10}$ & $1.23 \times 10^{10}$ & $4.84 \times 10^{10}$ \\
        Qwen3 MoE & $7.89 \times 10^8$ & $2.52 \times 10^{10}$ & $1.34 \times 10^{10}$ & $2.84 \times 10^{10}$ \\
        Qwen3 32B & $1.07 \times 10^9$ & $1.72 \times 10^{10}$ & $1.21 \times 10^{10}$ & $5.03 \times 10^{10}$ \\
        Llama 4 M & $1.01 \times 10^9$ & $8.05 \times 10^{9}$ & $6.04 \times 10^{9}$ & $2.42 \times 10^{10}$ \\
        MM M1 & $9.23 \times 10^8$ & $3.42 \times 10^9$ & $3.75 \times 10^{10}$ & $5.44 \times 10^{10}$ \\
        ERNIE 4.5 & $9.06 \times 10^8$ & $1.45 \times 10^{10}$ & $1.63 \times 10^{10}$ & $7.61 \times 10^{10}$ \\
        Pangu Pro & $8.05 \times 10^8$ & $8.05 \times 10^9$ & $6.04 \times 10^9$ & $2.38 \times 10^{10}$ \\
        \sys & $2.56 \times 10^8$ & $3.27 \times 10^{10}$ & $2.07 \times 10^{10}$ & $5.33 \times 10^{10}$ \\
        \bottomrule
    \end{tabular}
    \caption{Theoretical computation and memory access per decoding token at 8K context length.}
    \label{tab:decoding_8k}
\end{table*}

\begin{table*}[h]
    \centering
    \begin{tabular}{l|p{2.7cm}|p{3.5cm}|p{3cm}|p{2.5cm}}
        \toprule
        Model & KV/State Memory Access (bytes) & Attention Computation w/o Linear (FLOPs) & Linear before and after Attention (FLOPs) & FFN Computation (FLOPs) \\
        \midrule
        DSv3 & $1.15 \times 10^9$ & $5.89 \times 10^{11}$ & $2.28 \times 10^{10}$ & $4.84 \times 10^{10}$ \\
        Kimi K2 & $1.15 \times 10^9$ & $2.95 \times 10^{11}$ & $1.23 \times 10^{10}$ & $4.84 \times 10^{10}$ \\
        Qwen3 MoE & $3.15 \times 10^9$ & $1.01 \times 10^{11}$ & $1.34 \times 10^{10}$ & $2.84 \times 10^{10}$ \\
        Qwen3 32B & $4.29 \times 10^9$ & $6.87 \times 10^{10}$ & $1.21 \times 10^{10}$ & $5.03 \times 10^{10}$ \\
        Llama 4 M & $2.21 \times 10^9$ & $1.41 \times 10^{10}$ & $6.04 \times 10^{9}$ & $2.42 \times 10^{10}$ \\
        MM M1 & $1.93 \times 10^9$ & $1.15 \times 10^{10}$ & $3.75 \times 10^{10}$ & $5.44 \times 10^{10}$ \\
        ERNIE 4.5 & $3.62 \times 10^9$ & $5.80 \times 10^{10}$ & $1.63 \times 10^{10}$ & $7.61 \times 10^{10}$ \\
        Pangu Pro & $3.22 \times 10^9$ & $3.22 \times 10^{10}$ & $6.04 \times 10^{9}$ & $2.38 \times 10^{10}$ \\
        \sys & $1.02 \times 10^9$ & $1.31 \times 10^{11}$ & $2.07 \times 10^{10}$ & $5.33 \times 10^{10}$ \\
        \bottomrule
    \end{tabular}
    \caption{Theoretical computation and memory access per decoding token at 32K context length.}
    \label{tab:decoding_32k}
\end{table*}

\subsection{Theoretical Flops and Memory Access}
\label{subsec:flops_and_memory}

We begin by examining the overall memory access and computational operations required for decoding each token.

Given that various quantization methods directly impact memory access and the type of floating point computation, we select widely used quantized versions for each model:

\begin{itemize}[leftmargin=*]
  \item \textbf{MLA family:} The official implementation of DSv3 uses \emph{BF16} for attention, with other parts in \emph{FP8}. However, recognizing the existence of an \emph{FP8} quantized version of MLA within the open-source community, we adopt \emph{FP8} quantization for the whole model. The same quantization is applied to Kimi K2.

  \item \textbf{GQA family:} The official release of Qwen3 includes full \emph{FP8} quantization, which we will use. For other models like ERNIE 4.5 and Pangu Pro MoE, to align with Qwen3, we also use the same quantization. We believe the risk of losing model accuracy is low given our own experience with GQA models.

  \item \textbf{Hybrid models:} The official quantization of Llama 4 Maverick and MiniMax M1 is conservative, especially for attention. As hybrid attention model's quantization remains largely unexplored for us, we mostly follow the official setup, \emph{i.e.,} \emph{BF16} KV for full attention layers because they are critical for long context tasks. We use \emph{FP32} for MiniMax M1's Lightning Attention states, the same as its official setup. We give Llama 4 Maverick a favor for using FP8 for its chunked GQA attention, again based on our experience with GQA. For all the other parts we adopt the same aggressive \emph{FP8} quantization like all other evaluated models, to have a fair comparison.

  \item \textbf{\sys:} we have successfully quantized \sys to be a full \emph{FP8} model without losing model accuracy. So we use full \emph{FP8} quantization, which aligns with MLA and GQA faimly.
  
\end{itemize}

\vspace{+0.1in}
If the hardware does not support \emph{FP8} quantization, we assume the use of \emph{INT8} weights and \emph{INT8} KV cache instead of \emph{FP8}, so the memory access remains the same. The computation will be in \emph{BF16} or \emph{FP16}.

The results are listed in Table~\ref{tab:decoding_8k} and~\ref{tab:decoding_32k}. With the assumption of AFD, we divide the model costs into three parts: attention (without linear projection), the linear projection before and after attention, and FFN. For the first part, we consider the KV cache size and the computation simultaneously, since they grow linearly with batch size and context length.

For the linear projection before and after attention, we assume they can achieve compute-bound performance with sufficient batching. In this case, the memory access of the weights is amortized, and the costs will be determined by FLOPs. There is an exception where the $q/k/v\_proj$ of MLA and MFA may not be able to run in H800's compute-bound area, due to those parts not being TP-friendly and may not have a large enough batch size for H800. This means we slightly underestimate MLA and MFA costs on H800. However, this is a relatively small part of the total costs and specific to H800, so we omit it for simplicity.


For FFN, we focus only on the activated computation volume because, using AFD for \emph{not-too-sparse} MoE, sufficient batching can always be accumulated for FFN to reach high MFU and amortize the memory access for weights. Further details on MoE sparsity are discussed in the \S\ref{sec:codesign}. In the worst case, over-sparse models like DSv3, Kimi K2 and Llama 4 Marverick may see their FFN cost doubling or even tripling on H800 in real deployment. For now, we omit it for simplicity and give them a favor.

We also omit the embedding table and the final output linear layer since they consume relatively small ($<5\%$) memory access and computation for these models, and they are not too different across different models.


\subsection{Theoretical Decoding Cost in USD}
\label{subsec:decoding_cost}

Next, we can calculate the theoretical decoding costs of the models on different accelerators. Table~\ref{tab:accelerator_spec} shows the accelerator specifications and their estimated prices on public clouds. 

\begin{table*}[h]
    \centering
    \begin{tabular}{p{2.5cm}|p{2.5cm}|p{2cm}|p{2cm}|p{2.5cm}|p{3cm}}
        \toprule
        \makecell{Accelerator} & \makecell{Price per card \\ per hour (USD)} & \makecell{BF16/FP16 \\ FLOPs} & \makecell{FP8 FLOPs} & \makecell{Memory \\ bandwidth (B/s)} & \makecell{Compute-bandwidth \\ ratio (roofline)} \\
        \midrule
        NVIDIA H800 & 2 & $9.89 \times 10^{14}$ & $1.98 \times 10^{15}$ & $3.35 \times 10^{12}$ & 591 \\
        NVIDIA H20 & 0.8 & $1.48 \times 10^{14}$ & $2.96 \times 10^{14}$ & $4.00 \times 10^{12}$ & 74 \\
        NVIDIA A800 & 0.75 & $3.12 \times 10^{14}$ & N/A & $2.00 \times 10^{12}$ & 156 \\
        Ascend 910B & 0.67* & $2.80 \times 10^{14}$ & N/A & $1.60 \times 10^{12}$ & 175 \\
        \bottomrule
    \end{tabular}
    \caption{Comparison of accelerator specifications. *We do not have publicly available 910B pricing. We estimate its price proportionally based on its FLOPs and A800's. As far as we know, there are multiple versions of 910B. We show the weakest and (presumably) most affordable one that we know.}
    \label{tab:accelerator_spec}
\end{table*}

Suppose, in the theoretically ideal case, accelerators constantly at their peak FLOPs and maximum memory bandwidth, we derive the unit costs of a floating-point operation ($U_{FLOP}$) and a byte of memory access ($U_{byte}$), in Table~\ref{tab:unit_cost}.

\begin{table}[h]
    \centering
    \begin{tabular}{l|p{2.5cm}|p{2.5cm}}
        \toprule
        \makecell{Accelerator} & \makecell{Cost per FLOP }& \makecell{Cost per byte of \\ memory access} \\
        \midrule
        H800 & $2.80 \times 10^{-19}$ & $1.66 \times 10^{-16}$ \\
        H20 & $7.51 \times 10^{-19}$ & $5.56 \times 10^{-17}$ \\
        A800 & $6.68 \times 10^{-19}$ & $1.04 \times 10^{-16}$ \\
        910B & $6.65 \times 10^{-19}$ & $1.16 \times 10^{-16}$ \\
        \bottomrule
    \end{tabular}
    \caption{The unit cost of different accelerators assuming full utilization for the whole month. For FLOP costs, we consider FP8 for H800 and H20, BF16/FP16 for A800 and 910B.}
    \label{tab:unit_cost}
\end{table}

The theoretical cost of the attention part is the larger of the attention's core computation and memory access costs, plus the linear computation before and after:
\[
\max(FLOP_{Attn}U_{FLOP}, Byte_{KV}U_{byte}) + FLOP_{Linear}U_{FLOP}
\]

Assuming, with AFD, we can keep the FFN part in the compute-bound region, the theoretical cost of the FFN part is simply the computation cost $FLOP_{FFN}U_{FLOP}$.

\begin{table*}[h]
    \centering
    \begin{tabular}{|c|c|c|c|c|c|c|c|c|c|c|c|c|}
        \hline
        \multirow{2}{*}{Model} & \multicolumn{4}{c|}{Attention cost per 1M tokens (8k)} & \multicolumn{4}{c|}{Attention cost per 1M tokens (32k)} & \multicolumn{4}{c|}{FFN cost per 1M tokens} \\ \cline{2-13} 
         & H800 & H20 & A800 & 910B & H800 & H20 & A800 & 910B & H800 & H20 & A800 & 910B \\ \hline
        DSv3 & 0.054 & 0.128 & 0.114 & 0.113 & 0.197 & 0.460 & 0.409 & 0.407 & 0.014 & 0.036 & 0.032 & 0.032 \\ \hline
        Kimi K2 & 0.051 & 0.065 & 0.057 & 0.057 & 0.194 & 0.231 & 0.205 & 0.204 & 0.014 & 0.036 & 0.032 & 0.032 \\ \hline
        Qwen3 MoE & 0.135 & 0.054 & 0.091 & 0.101 & 0.527 & 0.185 & 0.338 & 0.376 & 0.008 & 0.021 & 0.019 & 0.019 \\ \hline
        Qwen3 32B & 0.181 & 0.069 & 0.120 & 0.133 & 0.716 & 0.248 & 0.455 & 0.508 & 0.014 & 0.038 & 0.034 & 0.033 \\ \hline
        Llama 4 M & 0.169 & 0.060 & 0.109 & 0.121 & 0.369 & 0.128 & 0.235 & 0.262 & \textbf{0.007} & \textbf{0.018} & \textbf{0.016} & \textbf{0.016} \\ \hline
        MM M1 & 0.164 & 0.079 & 0.121 & 0.132 & 0.330 & 0.135 & 0.226 & 0.249 & 0.015 & 0.041 & 0.036 & 0.036 \\ \hline
        ERNIE 4.5 & 0.155 & 0.063 & 0.105 & 0.116 & 0.606 & 0.214 & 0.388 & 0.432 & 0.021 & 0.057 & 0.051 & 0.051 \\ \hline
        Pangu Pro MoE & 0.135 & 0.049 & 0.088 & 0.098 & 0.536 & 0.183 & 0.340 & 0.379 & \textbf{0.007} & \textbf{0.018} & \textbf{0.016} & \textbf{0.016} \\ \hline
        \sys & \textbf{0.048} & \textbf{0.040} & \textbf{0.040} & \textbf{0.043} & \textbf{0.176} & \textbf{0.114} & \textbf{0.120} & \textbf{0.133} & 0.015 & 0.040 & 0.036 & 0.035 \\ \hline
    \end{tabular}
    \caption{Theoretical decoding cost analysis for each model on each hardware, in USD. As a reminder, these models have different number of activated parameters: DSv3 37B, Qwen3 MoE 22B, Qwen3 32B, MM M1 46B, ERNIE 4.5 47B, Pangu Pro MoE 16.5B and \sys 38B.}
    \label{tab:decoding_cost_full}
\end{table*}

Combining the attention and FFN parts, we obtain Table~\ref{tab:decoding_cost_full}. The final costs of different deployment choices can be directly computed. For example, we can add the attention and FFN parts together for different context on each hardware. For AFD, we choose the cheapest hardware for attention cost and FFN cost, respectively, and then sum them up. We assume all communication time on network can be overlapped by computation in a multi-batch pipeline, so the communication costs are ignored. 

\begin{figure*}[!t]
    \centering
    \includegraphics[width=0.49\linewidth]{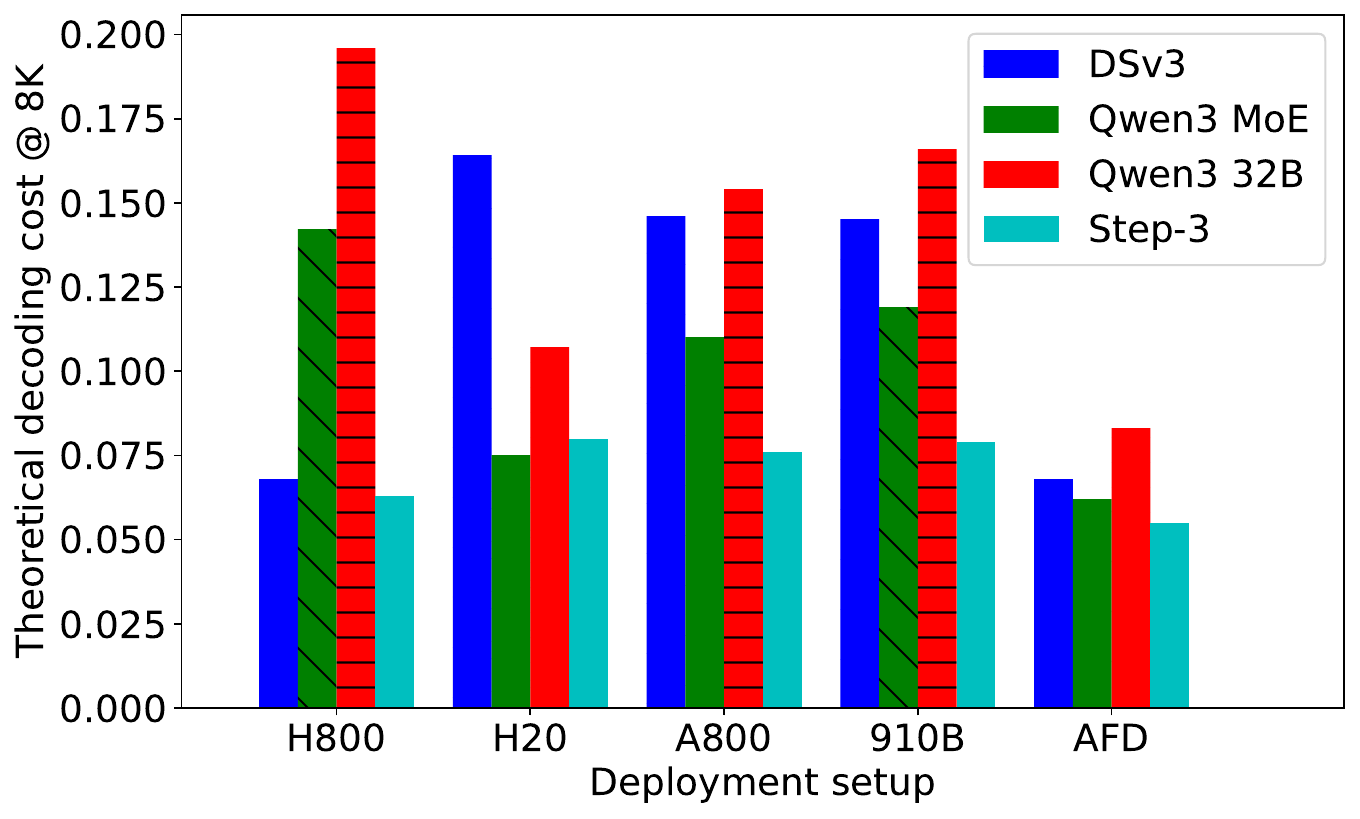}
    \includegraphics[width=0.49\linewidth]{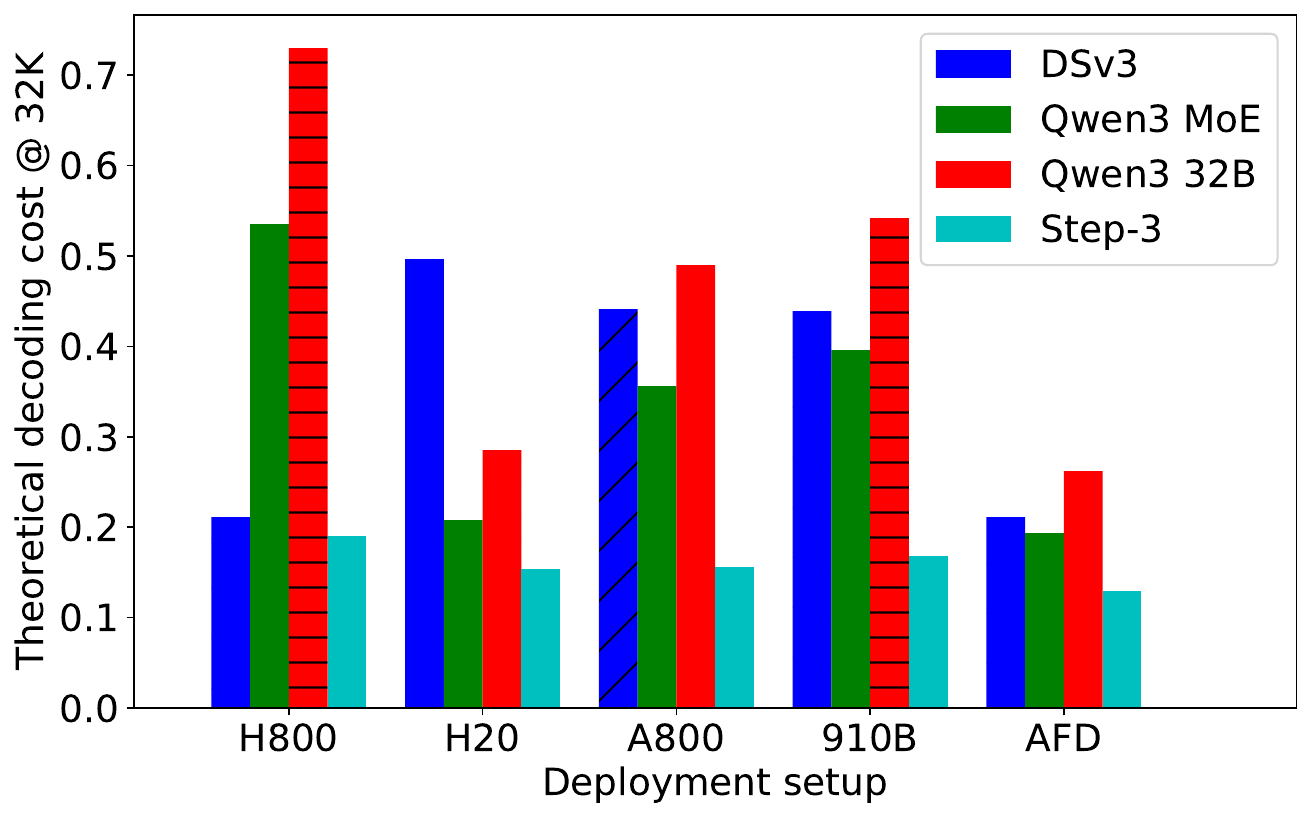}
    \caption{Decoding costs (per 1M tokens) of different models and inference configurations. For AFD, we combine the lowest costs on different hardware for attention and FFN, respectively. Reminder: \sys has the most activated parameters among them.}
    \label{fig:deployment_costs}
\end{figure*}

For brevity, we only show the results of Qwen family (representative for GQA models), DSv3 (representative for MLA models), and \sys in Figure~\ref{fig:deployment_costs}. With all the results shown, we make the following observations:

\parabf{Observation 1: \sys has the lowest decoding costs.}
When at 8K context length, \sys is the most cost-effective at $0.055$ per 1M decoding tokens (with AFD, H800 and H20), lower than DSv3's $0.068$ (with EP and H800) and Qwen MoE's $0.062$ (with AFD, H800 and H20). The advantage is larger at 32K context, with \sys at $0.129$, significantly lower than DSv3's $0.211$ and Qwen-3 MoE's $0.193$.

\parabf{Observation 2: Total and activated parameter numbers are bad indicator for decoding costs.} Qwen3 32B has much less total parameters than DSv3 and \sys, and also slightly less activated parameters. However, the decoding cost of Qwen3 32B is the highest among all models in Figure~\ref{fig:deployment_costs}.

\parabf{Observation 3: The cost of attention is dominating the total decoding cost.}
It is clear in Table~\ref{tab:decoding_cost_full}, at 8K context length, attention is already significantly more expensive than FFN. The gap grows quickly with longer context, given that FFN's cost is irrelevant to context length. This means the attention design matters much more than activated number of parameters -- that's the reason for Observation 2.

\parabf{Observation 4: Hardware friendliness.} DSv3's MLA is quite unfriendly to hardware other than H800, resulting in multi-fold increase when running on hardware weaker than H800. GQA models like Qwen3 are quite unfriendly to hardware other than H20, because of large KV sizes. In contrast, \sys's MFA is more hardware-friendly, with minimal cost differences for weaker hardware. We show this in Figure~\ref{fig:deployment_costs}.

\subsection{Demystifying Model Design Choices}
\label{subsec:demystify}

In this section, we discuss some ongoing model design trends in the community. We especially focus on the decoding phase. 

\parabf{Linear attention and hybrid models.} 
Linear attention is a promising direction but still faces challenges in long context tasks. A practical workaround is ``hybrid models'', which consist of two types of attention layers; most are linear attention, while the rest are traditional full attention. 
For example, MM M1, using a hybrid architecture with 70 layers of linear attention and 10 layers of GQA full attention, exhibits significantly slower KV growth with context length compared to full-GQA models like Qwen3. The design of Llama 4 Maverick is similar except for the layer numbers.

However, such hybrid models have two additional challenges for inference systems. 

\begin{figure}[!t]
    \centering
    \includegraphics[width=0.48\linewidth]{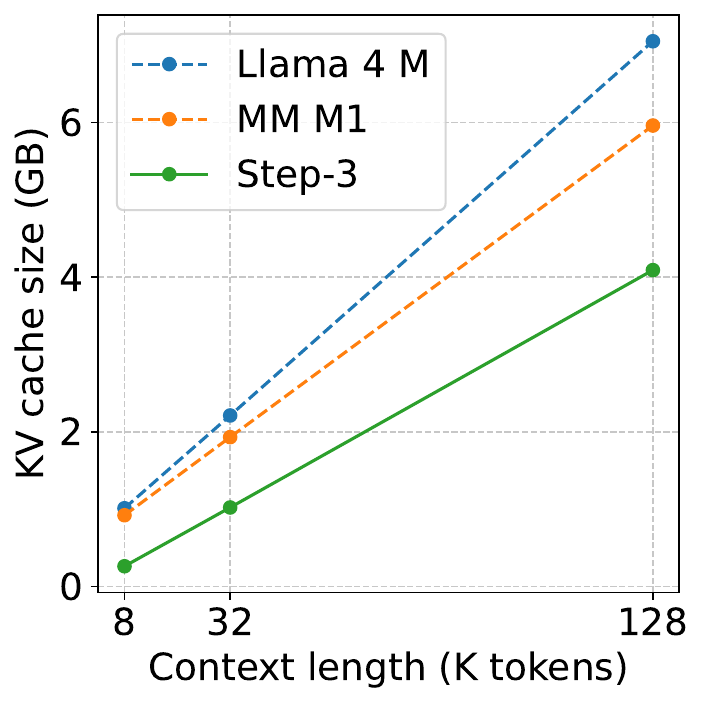}
    \includegraphics[width=0.5\linewidth]{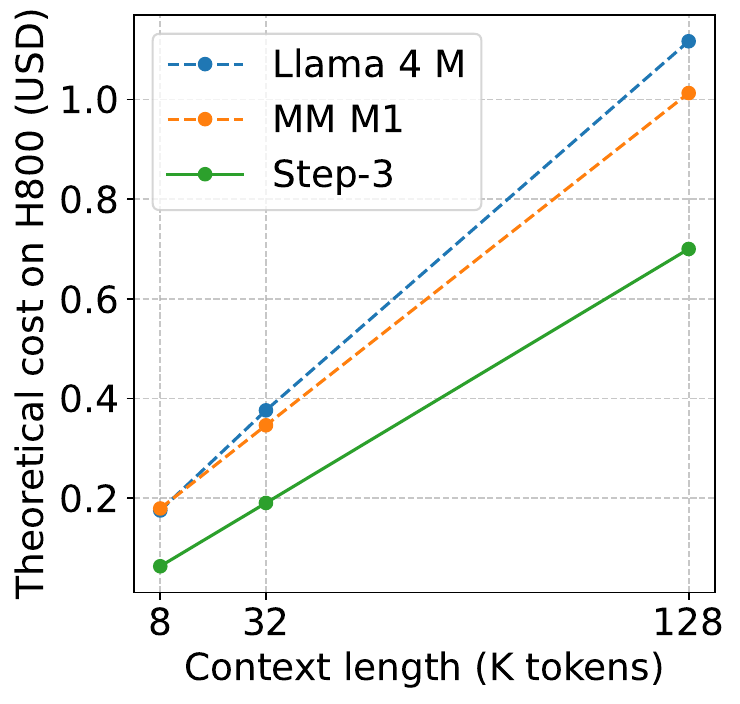}
    \caption{Total KV cache size and decoding cost comparison on H800 with hybrid linear attention models like MiniMax M1 and Llama 4 Maverick.}
    \label{fig:linear_attention}
\end{figure}

First, while the number of full attention layers seems small, they may still ruin the point of using linear attention for saving KV cache. MM M1 and Llama 4 Maverick's full attention part alone (based on the official quantization scheme) has a larger KV cache volume than \sys's entire model. No matter how much the rest of the linear attention layers save, no matter how long the context is, the total memory access will be larger than \sys, as shown in Figure~\ref{fig:linear_attention}.

Second, the time spent on each layer will be largely unbalanced -- when running with long context, the full GQA layers consume much more time than the linear attention layers. This may not be a problem for single-node inference deployment, but can be quite troublesome for distributed inference deployment (especially AFD) when one tries to build a pipeline to hide communication time. The imbalance of layer times can cause significant pipeline bubbles.

In Figure~\ref{fig:linear_attention}, we compare MM M1 and Llama 4 Maverick with \sys using a single hardware (H800) setup. Due to the reason above, they always have higher decoding costs than \sys despite most of their layers being linear attention. Admittedly, using hardware with cheaper memory bandwidth (like H20) can largely narrow the gap. But fundamentally, they require more KV cache access than \sys in the end.

We call for hybrid model designs that are more friendly to inference systems. One should design the full attention part carefully so that it does not ruin the cost saving from linear attention. Also, try to make every layer hybrid so that the time for each layer is balanced, instead of having a few slow layers that may limit the potentials of running in a distributed pipeline.

\parabf{``Hardware-optimized design'' -- for training or decoding?}
Designing a model that is optimized for a given hardware is not a new concept. In this paper, we include Pangu Pro MoE, a model claimed to be specifically optimized for Huawei's own accelerator, 910B.

\begin{figure}[!t]
    \centering
    \includegraphics[width=0.95\linewidth]{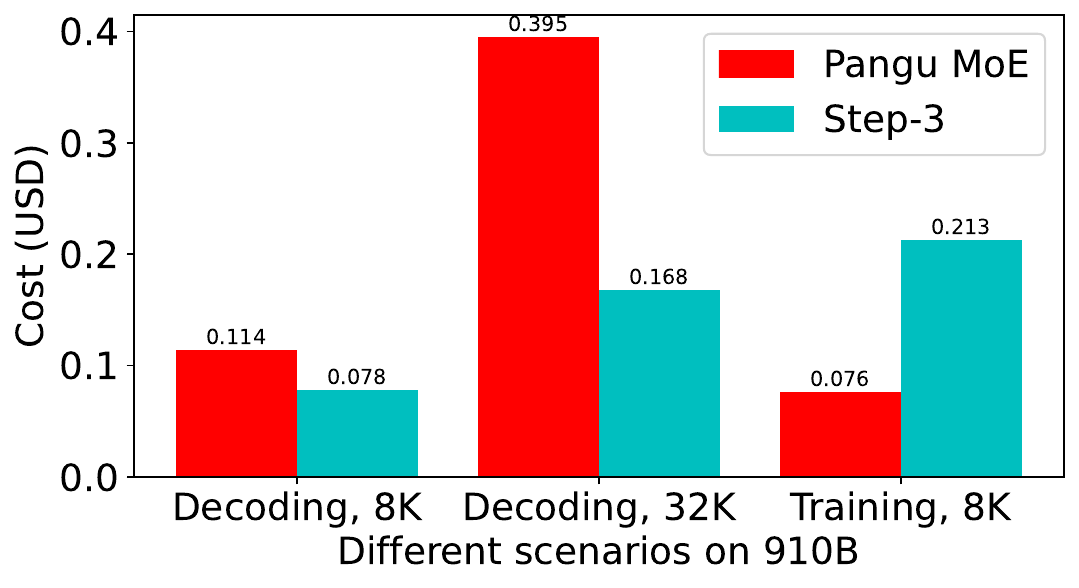}
    \caption{\sys and Pangu Pro MoE have very different trends of decoding cost and training cost.}
    \label{fig:pangu}
\end{figure}

However, in our analysis, the decoding cost of Pangu Pro MoE \emph{on 910B} is not low -- it is theoretically much larger than \sys (Figure~\ref{fig:pangu}). Remember, Pangu Pro MoE has only 16.5B activated parameters, less than half of \sys's! It is evident that Pangu Pro MoE's decoding on 910B is not cost-effective at all.

To be fair, the main focus of Pangu Pro MoE was not about decoding cost, it was about training. We also show a rough estimation of training cost per 1M token assuming 100\% MFU,\footnote{One can assume a more practical MFU like 40\%, but the trend will not reverse.} purely based on the theoretical FLOPs. We see that Pangu Pro MoE indeed is more than 50\% cheaper than \sys to train, reflecting the difference in activated parameters.

The lesson is, be clear about the goal during model-system co-design. Training and inference can be vastly different. Training costs are largely tied to the number of activated parameters, while lowering decoding costs requires additional model-system co-design. We will discuss the co-design points immediately.

\section{Model-System Co-design}
\label{sec:codesign}

\subsection{Matching Attention Arithmetic Intensity with Hardware}
\label{subsec:attention}

Readers paying attention (pun intended) may notice that in Tables~\ref{tab:decoding_8k} and~\ref{tab:decoding_32k}, \sys’s MFA exhibits only a ~10\% reduction in KV memory access volume compared with DSv3’s MLA. Yet in Table~\ref{tab:decoding_cost_full}, \sys’s attention cost is reduced by half or more in many cases. Why?
The result stems from the design of MFA. 

As pointed out in prior work~\cite{deepseek_nsa, gla}, each attention design has an inherent property called \emph{arithmetic intensity}. It is the ratio of the arithmetic operations needed for each byte of KV accessed from memory. Different batch sizes or context lengths do not change the arithmetic intensity.


The better the match between attention's arithmetic intensity and a hardware's ``computation-bandwidth ratio'' (or referred to as \emph{roofline}) (see Table~\ref{tab:accelerator_spec}), the more likely it is to achieve good efficiency on that hardware. Otherwise, significant bottlenecks may occur, either compute-bound or memory-bound. 

\begin{figure}[!t]
    \centering
    \includegraphics[width=0.95\linewidth]{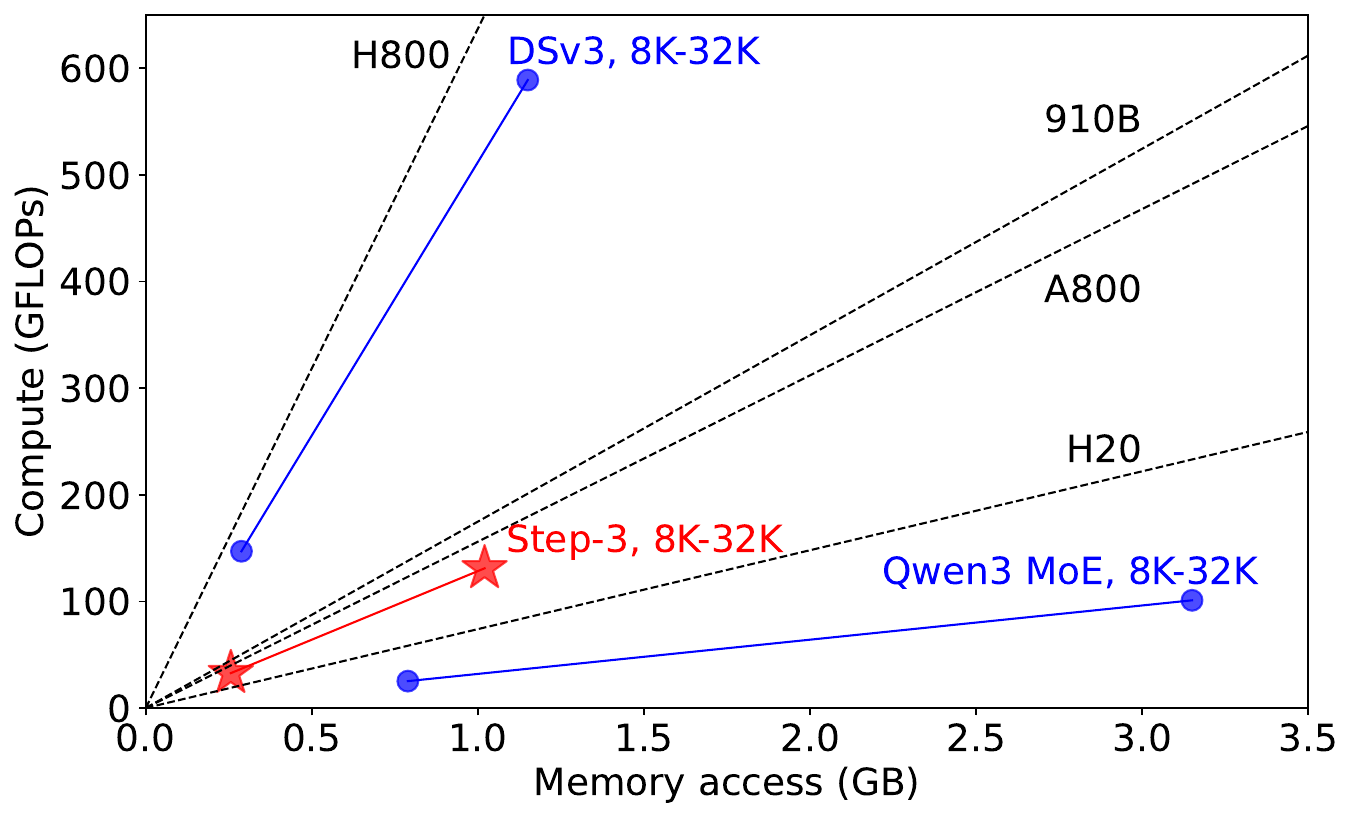}
    \caption{The compute and memory access of different attention designs during decoding, including DSv3's MLA, Qwen3 MoE's GQA and \sys's MFA. The compute-memory-bandwidth ratios of different hardware are also plotted.}
    \label{fig:arithmetic_intensity}
\end{figure}

With \sys's MFA design, its arithmetic intensity is 128 (assuming 8-bit quantization of KV). It is much closer to A800 (roofline is 156) and 910B (roofline is 175) than DSv3's MLA (arithmetic intensity is 512). On H20 (roofline is 74), \sys's gap is also not too large compared with Qwen3 MoE (arithmetic intensity is 32). To better illustrate, we show the above models and hardware in Figure~\ref{fig:arithmetic_intensity}. We show how compute and memory access grows with context length, from 8K to 32K, for each model. Correspondingly, we also plot a line for each hardware with the slope based on their computation-bandwidth ratio.

In Figure~\ref{fig:arithmetic_intensity}, it is also clear that \sys's MFA achieves low computation and memory access simultaneously. Namely, its required computation is one-fourth of DSv3's, and its required memory access is one-third of Qwen3's. This enables \sys to maintain low costs even on accelerators whose roofline does not match \sys well. 


\sys' MFA achieves the more balanced arithmetic intensity and low overhead without cutting corners. In fact, its attention effective rank~\cite{hu2025multimatrixfactorizationattention} is 16,384, the same as DSv3's MLA and larger than Qwen3 MoE's 8,192. 

\sys chooses slightly lower arithmetic intensity than most of the hardware's roofline, to leave room for future optimizations like quantization and MTP, as discussed next.



\subsection{Discussion: Quantization and MTP}
\label{subsec:MTP}

\parabf{Quantization:} All models can adopt more aggressive quantization strategies than those we have assumed. A particularly noteworthy quantization approach is \emph{low-bit storage with high-bit computation}, \emph{e.g.,} storing KV in 4-bit but performing attention calculation in 8-bit. This effectively \emph{doubles} the arithmetic intensity of each attention design. Such a change has different meanings for different attention designs. We still use DSv3, Qwen3, and \sys as examples:

\begin{itemize}[leftmargin=*]
    \item Implications for DSv3: Because DSv3's arithmetic intensity is already close to H800's roofline and much higher than other hardware, such quantization scheme will not improve efficiency.
    \item Implications for Qwen3: It might enable GQA-family models to get closer to or surpass H20's roofline. It can benefit on all hardware listed.
    \item Implications for \sys: This could potentially turn arithmetic intensity to exceed the roofline of A800 and 910B, but still not far off. There should be moderate performance gain. It may benefit a lot on H800 with higher roofline.
\end{itemize}


For quantization schemes that use the same format for KV storage and attention computation (assuming the hardware has native support), we anticipate that those will not significantly alter the overall trends of different models. 

Regarding hybrid models like MM M1, many (including ourselves) may wonder if aggressive KV quantization is feasible. However, given that there are only 8 layers of full attention and they might be more sensitive to quantized KV, we adopt a more conservative approach -- using the official setup -- in this paper. We look forward to more in-depth research on this topic.

\parabf{Multi-Token Prediction (MTP): }
MTP and the "low-bit storage, high-bit computation" quantization scheme have similar effects on arithmetic intensity -- \emph{doubling (or even multiplying)} it. Therefore, similar to the previous discussion, DSv3 is the least MTP-friendly model. GQA and MFA (\sys) models can leverage MTP to enhance throughput on various hardware.

However, MTP's impact is global -- enabling MTP also alters the computation load of FFN. Under the assumption of AFD, where FFN can always get enough batch to run with high MFU (see the next section), MTP could actually incur additional costs. MTP is not 100\% accurate in predicting additional tokens, yet FFN's cost is always increased regardless of prediction accuracy. One must be very careful in deciding whether to enable MTP.

\parabf{Summary:}
\sys's MFA design and its arithmetic intensity allows applying further KV quantization or enabling MTP to gain further cost savings than the results in Table~\ref{tab:decoding_cost_full}. In principle, Qwen3 and other GQA-based models could benefit from similar mechanisms. However, due to the high arithmetic intensity of its MLA, DSv3 may not see substantial benefits from further KV storage quantization or enabling MTP in large-batch, high-throughput scenarios.


\subsection{FFN's Batch Requirement for High MFU}
\label{subsec:batch_for_mfu}

Next, we discuss the costs of the Feed-Forward Network (FFN). The majority of FFN computation involves matrix multiplications, with a very small portion for activation functions. Most memory accesses are for model weights, with a very smaller portion for input and output hidden features. For simplicity, we will focus on matrix multiplications and model weight accesses.

For the matrix multiplication in FFN computation, the number of floating-point operations (FLOPs) is given by:
\[
2 \times N_{\text{token}} \times W_{\text{FFN}}
\]
where \( N_{\text{token}} \) represents the number of tokens processed in a batched FFN computation. In decoding, it is equivalent to the batch size $B$ entering the FFN (without MTP). \( W_{\text{FFN}} \) denotes the number of model weights in the FFN. Clearly, the computation-to-memory access ratio (assuming 8-bit weight storage) is $2 \times  N_{\text{token}}$, or $2 \times B$.

In the roofline model, to achieve good MFU, the computation-to-memory access ratio should at least match the hardware's roofline, as shown in Table~\ref{tab:accelerator_spec}. The corresponding ideal batch size, denoted as \( B_{\text{dense}} \), should at least be:
\[
2 \times B_{\text{dense}} \ge \frac{\text{FLOPs}}{\text{Bandwidth}}
\]

With a batch size that activates all experts, MoE increases the proportion between memory accesses and computation. We define the sparsity of MoE as \( S \). For example:
- If 2 experts are chosen from 8, then \( S = \frac{1}{4} \).
- If 8 experts are chosen from 256 plus one shared expert, then \( S = \frac{9}{256} \).

For MoE models, the ideal batch size for high MFU is:
\[
B_{\text{MoE}} = \frac{B_{\text{dense}}}{S}
\]
which can be several to tens of times larger than in dense models. Combined with the above equations, we get:
\[
B_{\text{MoE}} \ge \frac{\text{FLOPs}}{2 \times S \times \text{Bandwidth}}
\]

\subsection{Optimal MoE Sparsity vs. Hardware}
\label{subsec:moe_sparsity}

For contemporary models with hundreds of billions of parameters and long sequence inference, the memory capacity of a single machine often cannot support the appropriate batch size, necessitating distributed deployment. Whether using EP deployment~\cite{deepep2025} or AFD in this paper, the hardware running FFN computations needs to receive input hidden features (with dimension \( H \)) via the network and transmit the FFN computation results back via the network. Assuming 8-bit precision dispatch and 16-bit precision combine, and a batch size that meets high MFU requirements, the total transmission volume is:
\[
3 \times H \times B_{\text{MoE}}
\]


With AFD and an ideal three-stage pipeline and the TPOT target of $50ms$, we need to keep the network communication time below $50\text{ms}/3=16.6\text{ms}$. We denote network bandwidth as \( \text{Net} \), distinguished from memory bandwidth $Bandwidth$, we get:

\[
\frac{3 \times H \times B_{\text{MoE}}}{\text{Net}} \le \frac{16.6\text{ms}}{L}
\]
where \( L \) is the number of model layers. Substituting the expression for \( B_{\text{MoE}} \), we obtain:
\[
\frac{H \times \text{FLOPs} \times L}{\text{Net} \times S \times \text{Bandwidth}} \le \frac{16.6\text{ms} \times 2}{3} = 11.1\text{ms}
\]

We can derive the "optimal MoE sparsity" acceptable by the hardware, referring to the sparsest MoE configuration that the hardware can support to achieve ideal MFU while perfectly hiding network communication:
\[
S \ge \frac{H \times \text{FLOPs} \times L}{\text{Net} \times \text{Bandwidth} \times 11.1\text{ms}}
\]

Next, we use \sys's MoE architecture as an example. Its hidden feature size is 7168 and the number of layers \( L \) is 61. Those numbers are identical to DSv3.
We substitute the hardware parameters for each accelerator. We assume H800 and H20 use $400Gbps \times 8$ NICs, while A800 and 910B use $200Gbps \times 8$ NICs\footnote{The maximum network bandwidth is determined by PCIe generations.}. Table~\ref{tab:sparsity} shows the results.

\begin{table}[h]
    \centering
    \begin{tabular}{|c|c|c|c|c|}
    \hline
    Accelerator & H800 & H20 & A800 & 910B \\ \hline
    Minimum $S$ & 0.058 & 0.007 & 0.031 & 0.034 \\ \hline
    \end{tabular}
    \caption{Minimum MoE sparsity for different hardware platforms to achieve good MFU, where $H=7168$, $L=61$.}
    \label{tab:sparsity}
\end{table}

It is clear that the optimal MoE sparsity varies significantly across hardware platforms. H20 can accommodate the sparsest MoE configuration due to its lower computational power and higher memory bandwidth, allowing it to achieve high MFU with a smaller batch size and better tolerate MoE sparsity. H800 is the least friendly to very sparse MoE. However, H800 has the most affordable unit cost per FLOP (Table~\ref{tab:unit_cost}).

To ensure that \sys can leverage high-roofline hardware like H800, we make sure \sys is not sparser than 0.058. In contrast, DSv3, for example, would require $(256+1) \times 0.058 -1=14$ MoE experts\footnote{The $+1$ and $-1$ in the formula is for shared expert.} to be activated to achieve good MFU on H800, which is much larger than the official 8 activated experts. In other words, if DSv3 activates more experts, the decoding costs may not rise much. It means it may be leaving extra model performance on the table.

Even worse, unideal hardware efficiency may exaggerate the problem. For example, on the H800 platform with DeepEP~\cite{deepep2025}, the measured average throughput per network card is 40GB/s instead of 50GB/s, which can lead to a 25\% increase in the optimal sparsity, \emph{e.g.,} 0.073 for H800. Considering all these, \sys chooses a sparsity of around 0.08 (including shared expert).

Being even sparser, Llama 4 Maverick and Kimi K2 will be even further from the high MFU region when running on H800.

To clarify, all the theoretical cost analysis in \S\ref{sec:cost} ignores the network bottleneck by assuming all network communication can be overlapped and all FFNs can run in high MFU states. The network bottleneck and MoE sparsity problems discussed in this section will further increase the actual costs of over-sparse models like DSv3, Kimi K2, and Llama 4 Maverick. 





\subsection{Discussion: Workaround for Over-sparsity}
\label{subsec:workaround}

The above analysis regarding sparsity \( S \) is based on AFD's deployment philosophy -- using just enough FFN instances and accumulating a large batch size for high MFU. In a relatively small TP or EP deployment, the MoE sparsity $S$ on each FFN instance is the same as the whole model. For example, two FFN instances running DSv3's 8-in-256 with $EP=2$ (in terms of servers) mean each instance runs 4-in-128. $S$ remains the same for each server.

However, there are workarounds that may increase $S$, to alleviate the network bottleneck, but at the cost of other aspects.


\parabf{Workaround 1: Large EP.}
When EP (in terms of servers) is sufficiently large, especially exceeding $K$ (the number of activated experts), the network traffic volume required by each FFN (or EP) server is reduced. This is the case for DSv3's official deployment that uses more than 10 servers as a giant EP deployment.

\parabf{Workaround 2: MoE Routing Restrictions.}
Limiting token routing to adjacent experts can also make each local portion of the model not as sparse as the entire model.

DSv3 employs both methods to mitigate the issue of its over-sparsity for the H800 platform. Kimi K2 follows DSv3 on Workaround 1, but removes Workaround 2. This may make the network bottleneck even worse than DSv3.

We must note that both approaches come with costs: 1) Workaround 1 is more susceptible to expert imbalance issues, reducing actual efficiency. 2) Workaround 2 adversely affects the model's expressiveness. The impact on model performance has not been well studied yet.

\sys's design avoids this sparsity issue, allowing it to use small TP, EP or TP + EP hybrid approaches during AFD. This minimizes the performance impact of expert imbalance and eliminates the need for any routing restrictions.

\section{Non-Flagship Hardware Support}
\label{sec:hardware}

With AFD, both the attention and FFN components can be easily scaled, respectively. This creates more opportunities to leverage non-flagship hardware for the attention part, or FFN part, or both.


For example, \sys's MFA workload running on H800 is memory bandwidth bound. It can be replaced by four L20s, on which MFA is still memory bandwidth bound. L20's memory bandwidth is more than 25\% of H800's, so in theory, with a 25\% batch size on each L20, four L20s can run as fast as an H800 in a DP manner. Thanks to AFD, we do not need to worry about the FFN part -- it can remain unchanged when we discuss the attention part. For network communication, an L20 server needs 25\% of bandwidth compared with an H800 server, \emph{i.e.,} $4\times200 Gbps$ vs. $8\times400 Gbps$. It is also easy to satisfy.


The main limitation is that, for both attention and FFN servers, weaker hardware must still meet the latency requirements for AFD's three or four-stage pipeline to meet SLA. For instance, a three-stage pipeline requires both attention and FFN computations to be kept within \( \frac{16.6\text{ms}}{61\text{ layers}} \approx 272\text{$\mu$s} \) for \sys. We illustrate this with L20.

\parabf{Attention:}
We consider the memory access requirement as a necessary condition for satisfying the latency requirement. One L20 can access \( 864\text{ GB/s} \times 272\text{$\mu$s} = 235\text{ MB} \) within 272$\mu$s. The linear parts\footnote{$o_{proj}$ uses TP=8, while other linear parts are replicated across 8 GPUs.} require memory access of 67 MB in total. Thus, kvcache cannot exceed \( 235 - 67 = 168\text{ MB} \). With each token's KV being 512 bytes, the total inference context length cannot exceed approximately 328K tokens. This means that if the average context length is 8K tokens, keeping the batch size below 41 is sufficient. The maximum context length for a single request is up to 328K, which is still reasonable. Of course, the hardware cannot always run at its peak memory bandwidth, and there is inter-GPU communication overhead we omit. But we think L20 in general is capable of running \sys's attention part.

However, using even weaker accelerators like L4 with a memory bandwidth of 300 GB/s would spend most of the 272$\mu$s timeframe just to access the linear part's 67 MB. Consequently, L4 is unlikely usable for \sys's attention. We recommend accelerators that are at least as powerful as L20.

Given that \sys's MFA has the smallest KV volume and moderate arithmetic intensity, it is relatively hardware-friendly for weaker hardware than other recent models with similar sizes.


\parabf{FFN:}
Similarly to attention, each FFN layer must complete within 272$\mu$s. Both computation and memory access must finish within this timeframe. Computation scales with batch size, and we aim to maximize batch size to push FFN into the compute-bound (high MFU) region. For convenience, we assume an appropriate batch size pushes FFN into the compute-bound region, utilizing only 50\% of the memory bandwidth. Real-world scenarios might vary, but we use this for illustration.

For an L20, this means it can support FFN up to \( 864\text{ GB/s} \times 50\% \times 272\text{$\mu$s} = 117\text{ MB} \).
For 61 layers in \sys, this totals 7.1 GB. There are eight L20s per server, which can accommodate 56.8 GB of FFN weights. For the size of \sys (around 300 GB FFN weights), we need \emph{six} L20 servers, or \emph{48} cards, to run in EP to meet the performance requirements. We consider this number reasonable, especially as it is still much smaller than DSv3's deployment.

Again, we consider a weaker card L4, whose memory bandwidth is only one third of L20. It means we need 144 cards to meet the FFN latency requirement for \sys. At this scale, we start to be concerned about other issues like expert imbalance, stability, etc.


The primary influencing factor here is the total number of FFN parameters. The larger the total parameters, the less friendly it is to weaker hardware. \sys strikes a good balance at the level of L20 cards.


\parabf{Summary:}
For models with hundreds of billions of parameters, using at least L20 or stronger cards is recommended. Stronger cards reduce the number of required FFN servers, benefiting system reliability and MoE load balancing.
\section{Implementation and Results}
\label{sec:impl}

\subsection{System Workflow and Optimizations}

\begin{figure}[t]
    \centering
    \includegraphics[width=0.8\linewidth]{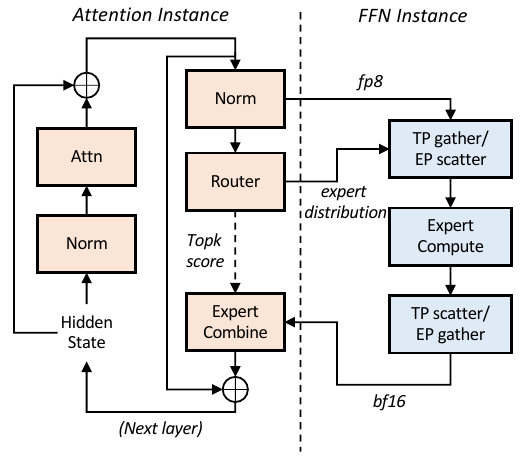}
    \caption{Module disaggregation in AFD architecture. FFN can be deployed in TP-only, EP-only, or a hybrid TP+EP way, depending on hardware and model architecture.}
    \label{fig:af_workflow}
\end{figure}

We describe our AFD system implementation details in this section. As shown in Figure~\ref{fig:af_workflow}, the AFD architecture is composed of two main components: 
(1) Attention instances: responsible for computing the attention modules, managing the KV cache, and performing the non-expert computation operations in MoE modules (e.g., routers). For \sys, we employ a local DP attention mechanism in which each GPU handles a batch of independent data.
(2) FFN instances: directly handle the pure MoE computation and multi-GPU communication necessary for TP or EP. Since FFN can be deployed in TP-only, or EP-only, or a hybrid TP+EP manner, the FFN instance is designed and implemented to be flexible and can be configured accordingly. 
We use TP-only FFN as an example, where the weights of all MoE experts are sharded in a tensor parallelism manner. As an FFN instance receives data from attention instances, it first performs an all-gather operation to collect the data from the TP region. After computation, it performs a reduce-scatter operation to aggregate and scatter the results back to the original GPUs, followed by the token transmission back to the attention instances.

The system can be configured to support multiple attention and FFN instances simultaneously.
During communication, the attention instances broadcast the \emph{FP8} tokens (quantized from the \emph{BF16} activation after the upstream normalization) to the FFN instances; conversely, the FFN instances return \emph{BF16} output to the attention instances to preserve high residual precision. For \sys, since the FFN instances spans multiple machines in a hybrid EP+TP way, the attention instances introduce a reduction module to combine all partial EP results from multiple FFN nodes.
In addition, the attention instances also need to transfer some small metadata, such as the expert distribution and \emph{FP8} tensor scale factors, to the FFN. The expert distribution is then used to dispatch the tokens and form an organized input for efficient expert computation. 
The metadata is typically small compared to the hidden state, and hence can be transferred with negligible overhead.

The design of our AFD system is simple, allowing for easy integration of different models and serving frameworks. For example, our attention instances are developed based on vLLM~\cite{vllm} with minimal changes, while FFN instances are implemented merely on top of a lightweight C++ communication library (will be introduced in \S\ref{subsec:impl_comm}) and simple PyTorch interfaces with no special dependencies.

\begin{figure}[t]
    \centering
    \includegraphics[width=\linewidth]{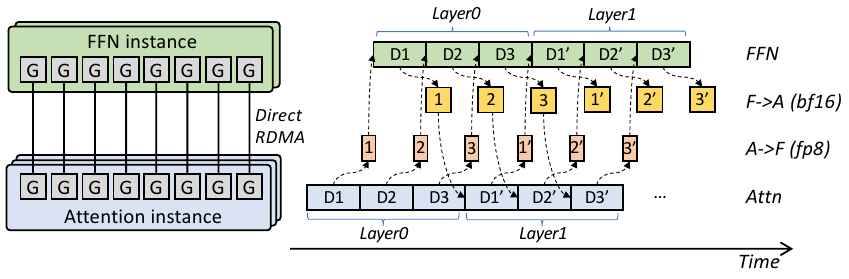}
    \caption{Communication topology and the multi-stages pipeline of the AFD architecture.}
    \label{fig:sys_architecture}
\end{figure}

\begin{figure*}[h!t]
\centering
\includegraphics[width=0.8\linewidth]{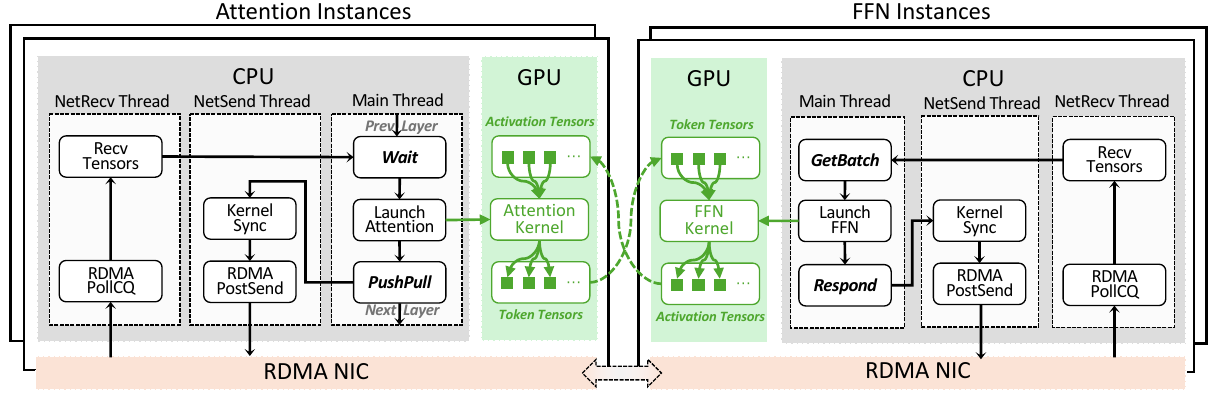}
\vspace{-4pt}
\caption{StepMesh communication workflow tailored for AFD.}
\label{fig:stepmesh:workflow}
\vspace{-10pt}
\end{figure*}

\begin{figure}[h!t]
\centering
\includegraphics[width=0.45\textwidth]{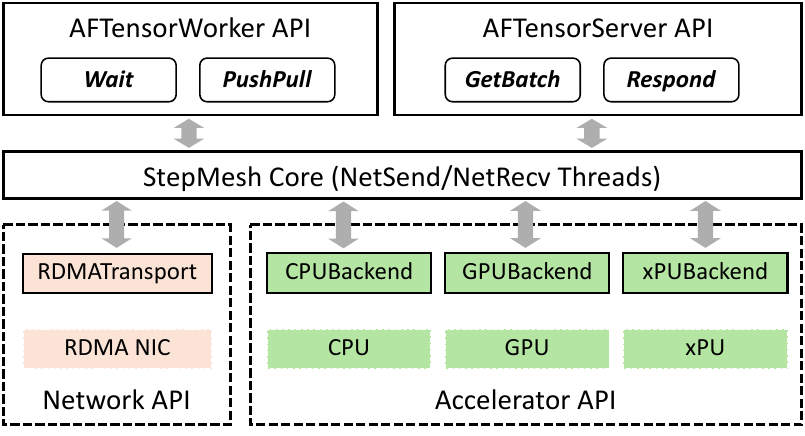}
\vspace{-4pt}
\caption{StepMesh framework for multiple accelerators. AFTensorWorker and AFTensorServer APIs are for attention and FFN instances respectively.}
\label{fig:stepmesh:framework}
\vspace{-0.15in}
\end{figure}

\parabf{Multi-stages Pipeline.}
\sys adopts a multi-stages pipeline to hide communication overhead and thus maximize overall throughput.
Figure~\ref{fig:sys_architecture} illustrates the data flow in the multi-stages pipeline. Starting from the attention instance, the system receives three input samples (D1, D2, D3). These samples are processed sequentially and then transmitted over the network to the FFN instance for computation. With careful workload orchestration, the computation time for each computation stage is made nearly identical, enabling efficient pipelining and minimizing idle periods.
The communication topology enables direct RDMA between GPUs, allowing data to be streamed in parallel with minimal latency, which can be easily hidden by the computation.
Note that the figure distinguishes A$\rightarrow$F and F$\rightarrow$A communication paths for simplicity. However, they represent two independent communication and do not compete for network bandwidth, allowing them to execute concurrently in practice.
As (D1, D2, D3) returns to the attention instance sequentially, the system can start processing the next layer in a streaming way, noted as (D1', D2', D3') in Figure~\ref{fig:sys_architecture}. This design allows the system to achieve high throughput while maintaining low latency, as the critical path does not delay the processing of each sample.

\parabf{Other Implementation Details.} 
We place the embedding and the LM head layers together with the attention instances since they incur small computation overhead. We develop tailored kernel optimizations for most kernels in the critical path, such as the \emph{FP8} GEMM and Flash Attention. For efficient NVLink communication for TP or EP within a single node, we leverage the NVLS APIs to implement the all-gather and reduce-scatter operations that not only can saturate NVLink bandwidth, but also can significantly reduce the GPU SM usage (particularly, our all-gather op is SM-free). The low SM usage is crucial for efficient communication-computation overlap, as revealed in previous work~\cite{flux,sigcomm25_disttrain}.

\subsection{StepMesh: AFD Communication Library}
\label{subsec:impl_comm}

AFD presents stringent performance challenges for communication libraries. For a 3-stage pipeline, AFD demands to complete transmission of \emph{FP8} tokens, scales, expert distribution, and \emph{BF16} activation between all attention and FFN instances within 272 $\mu$s (\S\ref{sec:hardware}). Existing communication libraries struggle to consistently meet the requirement. Moreover, current libraries like NCCL and DeepEP introduce additional GPU SM usage dedicated to communication, inherently compromising the computation speed of attention and FFN. AFD also introduces a novel communication pattern, distinct from existing collectives and not well-supported. While workarounds like ncclSend/ncclRecv can be employed, they inevitably sacrifice performance. Addressing these challenges, we develop \emph{StepMesh}, a specialized communication library for AFD based on GPUDirect RDMA, offering ultra-low latency, zero SM usage, and flexible communication.

\parabf{Communication Workflow Tailored for AFD Pipelines.}
Figure~\ref{fig:stepmesh:workflow} illustrates the design choices of StepMesh to optimally align with the AFD pipeline stages.
1) Asynchronous APIs and dedicated threads: StepMesh offers asynchronous APIs and utilizes independent threads for network receiving and sending. The CPU latency of each thread is meticulously designed to meet stringent latency requirements, ensuring smooth and efficient data flow.
2) CPU-Based operation execution: To avoid contention for GPU SM resources with computation threads, StepMesh executes all communication operations—such as RDMA PostSend—on CPUs. It leverages NUMA-aware CPU core binding to minimize processing jitters and ensure stable performance. However, we continue to observe certain jitters originating from GPU APIs, such as the GPU kernel synchronization API (cudaEventSync). In future iterations, we plan to explore IBGDA~\cite{ibgda} to eliminate GPU kernel synchronization on CPUs, thereby further reducing communication latency.
3) Pre-registered tensors for efficient communication: StepMesh supports direct memory transmission for GPU tensors, eliminating the need for serialization/deserialization or memory copying. StepMesh requires users to register tensors, identified by unique tensor keys, before initiating communication. This registration process is flexible and can remove some time-consuming operations. For instance, FFN does not need to concatenate tensors from different attention instances. Instead, these tensors can be directly sliced from contiguous GPU memory that has been pre-registered, streamlining the communication process and improving efficiency.


\parabf{Support Heterogeneous Accelerators.}
Figure~\ref{fig:stepmesh:framework} presents the StepMesh framework, designed to be highly extensible and capable of integrating new types of accelerators.
This framework treats accelerators as backends and establishes a set of backend interfaces that are crucial for AFD communication. These interfaces encompass essential functionalities such as memory allocation and stream synchronization. By adhering to these well-defined interfaces, new accelerators can be effortlessly integrated into the StepMesh framework. This streamlined and future-proof integration process allows for the rapid adoption of emerging hardware technologies, ensuring that the system remains at the cutting edge of performance and efficiency. StepMesh enables seamless communication between heterogeneous accelerators, fostering an environment where different types of hardware can collaborate effectively. This capability is essential for building cost-effective AFD systems that leverage a mix of accelerators to achieve optimal performance and resource utilization.


\parabf{Co-evolution with Networks.}
Our AFD system operates on a Rail-Optimized RoCE network. The following optimizations have been implemented for deploying AFD over RoCE.
1) Topology-aware deployment: Attention and FFN instances are strategically connected to the same Top-of-Rack (ToR) switches. This deployment ensures that communication between any attention and FFN instance experiences uniform network latency, resulting in balanced communication costs and mitigating straggling issues, where certain nodes lag behind others, causing bottlenecks.
2) PFC-Only Transport: We disable congestion control and rely solely on ToR-NIC Priority Flow Control (PFC). PFC maintains a lossless network environment, crucial for the high-performance and low-latency requirements of AFD pipelines.
3) Balancing traffic between NIC ports: In our network, each GPU connects to the network through two NIC ports configured with link aggregation. To fully leverage the available bandwidth, for every communication pair (e.g., between an attention and FFN instance), we establish two RDMA Queue Pairs and assign them to the respective ports. This setup effectively balances traffic across both ports, optimizing data transmission efficiency and ensuring that the combined bandwidth is utilized effectively.

StepMesh is developed based on~\cite{byteps}, and we also make it available as an open-source project. Interested developers can access, contribute to, and utilize the library by visiting~\url{https://github.com/stepfun-ai/StepMesh}.

\subsection{Performance Results} 
\label{subsec:perf}

\begin{table}[t]
    \centering
    \begin{tabular}{|c|c|c|c|c|}
    \hline
    {Model} & \makecell[t]{Context\\Len (avg)} & \makecell[t]{\# Hopper \\ GPUs} & \makecell[c]{Peak TGS}  \\ \hline
    DSv3-blog~\cite{dsv3_blog} & 4989 & 144 & 1850 \\ \hline
    DSv3-profile~\cite{ds_profile} & 4096 & 128 & 2324 \\ \hline
    \makecell[t]{\sys\\(\emph{BF16} attention)} & {4096} & \makecell[t]{40\\(3A2F)} & 3321 \\ \hline
    \makecell[t]{\sys\\(\emph{FP8} attention)} & {4096} & \makecell[t]{32\\(2A2F)} & 4039 \\ \hline
    \makecell[t]{\sys\\(\emph{FP8} attention)} & {8192} & \makecell[t]{48\\(4A2F)} & 2643 \\ \hline
    \end{tabular}
    \caption{Performance comparison with reported number of DSv3 under 20 tokens/s decoding SLA. TGS: Tokens/GPU/s.}
    \label{tab:e2e_perf}
\end{table}

\parabf{End-to-End Performance.}
We compare \sys with DSv3, since it proposed the most representative distributed inference solution. Its official blog reports sustained average decoding throughput of 1,850 tokens/GPU/s (TGS) on H800, with 4,989 context on average. A higher peak performance in profiling is reported in~\cite{ds_profile}, at 2,324 TGS with 4,096 context length on H800. Both numbers are obtained under 20 tokens/s decoding SLA.\footnote{We are also aware of higher numbers like~\cite{sglang_ds}. However, we do not compare with them because they do not run with the same 20 tokens/s decoding SLA or have shorter context length.}

To have a direct comparison, we also test \sys's decoding with 4,096 average context length on latest Hopper GPUs. GEMM runs in \emph{FP8} precision. While adhering to 20 tokens/s decoding SLA, \sys achieves 3,910 TGS on long-term average and 4,039 TGS (with \emph{FP8} attention) in a peak minute, around 74\% higher than DSv3.
We summarize the results in Table~\ref{tab:e2e_perf}. 
We acknowledge that there is room for further improving DSv3 with more quantization, kernel optimizations, or better Hopper GPUs. However, we are confident that with the same level of optimizations and hardware, \sys can still achieve significantly higher throughput than DSv3.

We are still working on a few implementation details to reduce jitter and bring the average throughput closer to the peak throughput. Those numbers are obtained \emph{without} MTP. As \S\ref{subsec:MTP} explained, \sys can benefit from MTP significantly on accelerators other than H20. A rough estimate is a 50\% (or more, for longer context) improvement, given that the attention efficiency can double with MTP while FFN remains the same (MFU is already high without MTP).

For the above 4K context length case, we use ``2A2F'' deployment, which means two attention instances plus two FFN instances, in total 32 GPUs. The total batch size is 6144, divided into three micro batches of 2,048 to fill the 3-stage pipeline. For different average context lengths, we can simply scale attention instances. For example, for 8K average context length, we can use ``4A2F'' and keep the same total batch size as 6144. The latency and MFU for each component and the total network traffic will remain the same, so the SLA still holds and total throughput remains the same. The peak TGS will fall to around $4039\times(2+2)/(4+2)=2693$. Readers can extrapolate the deployment solution and performance numbers for longer context, \emph{e.g.,} ``16A2F'' for average 32K context length with 898 TGS, etc.

Note that the above scenarios are where \sys has the least advantage in cost saving compared with DSv3 using EP deployment. \sys's advantage will widen with longer context and on cheaper hardware than H800 (\S\ref{sec:cost}).

\parabf{Ablation: Attention Quantization.}
Our previous results on \sys are with \emph{FP8} attention. We also test \emph{BF16} attention to understand the gain from quantization. Since the attention cost increases, we use ``3A2F'' with a total batch size of 6048, close to the previous 6144. Each attention instance then processes 6048/3/3=672 samples for each micro batch. As shown in Table~\ref{tab:e2e_perf}, the result is 3,321 TGS, around 18\% lower than \emph{FP8} attention. But it still outperforms DSv3 by a large margin.

\parabf{Ablation: MFA.}
To further understand the performance gain, we conduct an ablation study on the attention layer of \sys, DSv3 and Qwen3-235B. They represent three different attention designs -- MFA, MLA and GQA, respectively. Since we only test the attention layer, the number also indicates the performance of an attention instance in real AFD deployment. As shown in Table~\ref{tab:attn_ablation}, MFA-Step3 achieves the lowest latency, followed by MLA-DSv3 and GQA-Qwen3. The performance gap is widened on H20 and A800, indicating that MFA is more efficient on lower-end accelerators. Also, the gap is larger on longer context lengths, which aligns with our analysis in \S\ref{sec:codesign}.

\begin{table}[t]
\centering
\begin{tabular}{|c|c|ccc|}
\hline
    \makecell[t]{Context\\Length} & \makecell[t]{Attention\\Type} & \multicolumn{3}{c|}{\makecell[c]{Time Per \\ Attention Layer (us)}} \\ \cline{3-5}
    & & H800 & H20 & A800  \\ \hline
    \multirow{3}{*}{8k} & MFA-Step3 & 281 & 438 & 531 \\ \cline{2-5}
                        & MLA-DSv3 & 372    & 1252    & -    \\ \cline{2-5}
                        & GQA-Qwen3 & 382 & 812 & 791 \\ \hline
    \multirow{3}{*}{32k} & MFA-Step3 & 791 & 1452 & 1484 \\ \cline{2-5}
                        & MLA-DSv3 & 1125    & 4817    & -    \\ \cline{2-5}
                        & GQA-Qwen3 & 1391 &  3042 & 3010 \\ \hline
\end{tabular}
\caption{Performance comparison of MFA/MLA/GQA. For MLA, we use FlashMLA which does not have official SM80 implementation, so its A800 number is not tested. We use FA3 (SM90) and FA2 (SM80) for MFA/GQA. Here the attention layer includes the linear projection before and after the core attention op. Each experiment uses 4 GPUs and a total batch size of 256. Both MFA and MLA use DP attention, while GQA uses TP attention. GEMM runs with \emph{FP8} (SM90) or \emph{INT8} (SM80) while attention runs with \emph{BF16}.}
\label{tab:attn_ablation}
\end{table}

\parabf{Ablation: Scaling \sys to > 600B.} Readers may wonder how much \sys's advantage is due to having fewer total parameters than DSv3. We consider the case to upcycle~\cite{upcycle_google, upcycle_nvidia} \sys's MoE FFN into the 600B parameters region, a similar size to DSv3. Since FFN is doubled, we will need ``4F'' instead of ``2F'' to keep per-token latency the same. However, suppose we do not increase activated parameters per token, upcycled \sys will have the same over-sparse problem as DSv3 and face network bandwidth limit. Calculation shows the $400Gbps \times 8$ network can only sustain a micro batch of 3,072 (8-bit dispatch, 16-bit combine) for each FFN instance. Thus, the final solution is ``3A4F'' running three micro batches of 3,072. Each A and F has the same or less load than the original \sys, so the 50ms TPOT SLA still holds. In this case, the TGS is 3,291. It shows the impact of over-sparsity (\S\ref{subsec:moe_sparsity}), compared with the original \sys's 4,039. Nevertheless, it is still much higher than DSv3's 2,324 with DeepEP. If we further align with the official DSv3 on running attention with BF16, based on profiling we estimate such upcycled \sys will run at around 2,880 TGS -- it shows the advantage of AFD over pure EP.


\section{Conclusion and Future Work}
\label{sec:conclusion}

This paper presents \sys, and how its model-system co-design achieves state-of-the-art level of decoding efficiency among LLMs of similar sizes. Meanwhile, we also explain how we leverage AFD for analysis and realize \sys's potentials. The immediate next step for us is to enable MTP and evaluate its performance gain for decoding. In the future, we will work on exploring new attention variants that continue to push the Pareto frontier of model volume and system costs. We also analyzed that today's interconnect limits the sparsity of MoE FFN if the goal is efficient decoding. To mitigate this problem, we are working with hardware vendors on novel high bandwidth domain designs~\cite{sigcomm25_infinitehbd}. With appropriate interconnect, we will pursue more sparsity for FFN.

\balance
\bibliographystyle{plain}
\bibliography{paper.bib}

\newpage
\appendix

\onecolumn

\noindent\textbf{\textit{All author lists are in alphabetical order.}}

\section*{Core System Contributors}

\begin{multicols}{4} 
\noindent
Bin Wang \\
Bojun Wang \\
Changyi Wan \\
Guanzhe Huang \\
Hanpeng Hu \\
Haonan Jia \\
Hao Nie \\
Mingliang Li \\
Nuo Chen \\
Siyu Chen \\
Song Yuan \\
Wuxun Xie \\
Xiaoniu Song \\
Xing Chen \\
Xingping Yang \\
Xuelin Zhang \\
Yanbo Yu \\
Yaoyu Wang \\
Yibo Zhu \\
Yimin Jiang \\
Yu Zhou \\
Yuanwei Lu
\end{multicols}

\section*{Core Model Architecture Contributors}

\begin{multicols}{4} 
\noindent
Houyi Li \\
Jingcheng Hu \\
Ka Man Lo \\
\end{multicols}

\section*{Contributors (Pretrain, Post-train, Multi-modal, System, Data)}

\begin{multicols}{4} 
\noindent
Ailin Huang	\\
Binxing Jiao	\\
Bo Li	\\
Boyu Chen	\\
Changxin Miao	\\
Chao Lou \\
Chen Hu	\\
Chen Xu	\\
Chenfeng Yu	\\
Chengyuan Yao	\\
Daokuan Lv	\\
Dapeng Shi	\\
Deshan Sun	\\
Ding Huang	\\
Dingyuan Hu	\\
Dongqing Pang 	\\
Enle Liu	\\
Fajie Zhang 	\\
Fanqi Wan	\\
Gulin Yan	\\
Han Zhang	\\
Han Zhou	\\
Hanghao Wu	\\
Hangyu Guo	\\
Hanqi Chen 	\\
Hanshan Zhang \\
Hao Wu 	\\
Haocheng Zhang	\\
Haolong Yan \\
Haoran Lv	\\
Haoran Wei	\\
Hebin Zhou	\\
Heng Wang	\\
Heng Wang	\\
Hongxin Li	\\
Hongyu Zhou	\\
Hongyuan Wang	\\
Huiyong Guo	\\
Jia Wang	\\
Jiahao Gong	\\
Jialing Xie	\\
Jian Zhou	\\
Jianjian Sun	\\
Jiaoren Wu	\\
Jiaran Zhang	\\
Jiayu Liu 	\\
Jie Cheng	\\
Jie Luo 	\\
Jie Yan	\\
Jie Yang	\\
Jieyi Hou	\\
Jinguang Zhang	\\
Jinlan Cao 	\\
Jisheng Yin	\\
Junfeng Liu	\\
Junhao Huang 	\\
Junzhe Lin	\\
Kaijun Tan	\\
Kaixiang Li	\\
Kang An	\\
Kangheng Lin	\\
Kenkun Liu	\\
Lei Yang	\\
Liang Zhao	\\
Liangyu Chen	\\
Lieyu Shi	\\
Liguo Tan 	\\
Lin Lin	\\
Lin Zhang 	\\
Lina Chen	\\
Liwen Huang 	\\
Liying Shi 	\\
Longlong Gu	\\
Mei Chen 	\\
Mengqiang Ren	\\
Ming Li	\\
Mingzhe Chen	\\
Na Wang 	\\
Nan Wu 	\\
Qi Han	\\
Qian Zhao	\\
Qiang Zhang	\\
Qianni Liu 	\\
Qiaohui Chen 	\\
Qiling Wu	\\
Qinglin He	\\
Qinyuan Tan 	\\
Qiufeng Wang \\
Qiuping Wu 	\\
Qiuyan Liang 	\\
Quan Sun	\\
Rui Li	\\
Ruihang Miao	\\
Ruosi Wan	\\
Ruyan Guo	\\
Shangwu Zhong \\
Shaoliang Pang	\\
Shengjie Fan	\\
Shijie Shang	\\
Shilei Jiang	\\
Shiliang Yang	\\
Shiming Hao	\\
Shuli Gao	\\
Siming Huang \\
Siqi Liu	\\
Tiancheng Cao	\\
Tianhao Cheng	\\
Tianhao Peng	\\
Wang You	\\
Wei Ji	\\
Wen Sun 	\\
Wenjin Deng	\\
Wenqing He	\\
Wenzhen Zheng	\\
Xi Chen	\\
Xiangwen Kong	\\
Xianzhen Luo	\\
Xiaobo Yang	\\
Xiaojia Liu	\\
Xiaoxiao Ren	\\
Xin Han	\\
Xin Li	\\
Xin Wu	\\
Xu Zhao	\\
Yanan Wei	\\
Yang Li 	\\
Yangguang Li	\\
Yangshijie Xu	\\
Yanming Xu	\\
Yaqiang Shi	\\
Yeqing Shen \\
Yi Yang	\\
Yifei Yang 	\\
Yifeng Gong 	\\
Yihan Chen	\\
Yijing Yang 	\\
Yinmin Zhang 	\\
Yizhuang Zhou	\\
Yuanhao Ding	\\
Yuantao Fan	\\
Yuanzhen Yang	\\
Yuchu Luo	\\
Yue Peng	\\
Yufan Lu	\\
Yuhang Deng 	\\
Yuhe Yin 	\\
Yujie Liu	\\
Yukun Chen	\\
Yuling Zhao 	\\
Yun Mou	\\
Yunlong Li 	\\
Yunzhou Ju	\\
Yusheng Li	\\
Yuxiang Yang	\\
Yuxiang Zhang	\\
Yuyang Chen	\\
Zejia Weng	\\
Zhe Xie	\\
Zheng Ge	\\
Zheng Gong	\\
Zhenyi Lu	\\
Zhewei Huang	\\
Zhichao Chang	\\
Zhiguo Huang	\\
Zhirui Wang \\
Zidong Yang	\\
Zili Wang	\\
Ziqi Wang \\
Zixin Zhang	\\
\end{multicols}

\section*{Sponsors}

\begin{multicols}{4} 
\noindent
Binxing Jiao \\
Daxin Jiang \\ 
Heung-Yeung Shum \\  
Xiangyu Zhang \\ 
Yibo Zhu
\end{multicols}

\end{document}